\documentclass[10pt,twocolumn,letterpaper]{article}

\usepackage[pagenumbers]{wacv} 
\usepackage{graphicx}
\usepackage{amsmath}
\usepackage{amssymb}
\usepackage{booktabs}
\usepackage{wrapfig}
\usepackage{caption}
\usepackage{hyperref}
\usepackage[accsupp]{axessibility}
\usepackage[capitalize]{cleveref}
\crefname{section}{Sec.}{Secs.}
\Crefname{section}{Section}{Sections}
\Crefname{table}{Table}{Tables}
\crefname{table}{Tab.}{Tabs.}

\def\wacvPaperID{2351} 
\def\confName{WACV}
\def\confYear{2025}

\begin{document}

\title{A Recipe for Geometry-Aware 3D Mesh Transformers}

\author{Mohammad Farazi\\
Arizona State University\\
Tempe, Arizona\\
{\tt\small mfarazi@asu.edu}
\and
Yalin Wang\\
Arizona State University\\
Tempe, Arizona\\
{\tt\small ylwang@asu.edu}
}
\maketitle

\begin{abstract}
Utilizing patch-based transformers for unstructured geometric data such as polygon meshes presents significant challenges, primarily due to the absence of a canonical ordering and variations in input sizes. Prior approaches to handling 3D meshes and point clouds have either relied on computationally intensive node-level tokens for large objects or resorted to resampling to standardize patch size. Moreover, these methods generally lack a geometry-aware, stable Structural Embedding (SE), often depending on simplistic absolute SEs such as 3D coordinates, which compromise isometry invariance essential for tasks like semantic segmentation. In our study, we meticulously examine the various components of a geometry-aware 3D mesh transformer, from tokenization to structural encoding, assessing the contribution of each. Initially, we introduce a spectral-preserving tokenization rooted in algebraic multigrid methods. Subsequently, we detail an approach for embedding features at the patch level, accommodating patches with variable node counts. Through comparative analyses against a baseline model employing simple point-wise Multi-Layer Perceptrons (MLP), our research highlights critical insights: 1) the importance of structural and positional embeddings facilitated by heat diffusion in general 3D mesh transformers; 2) the effectiveness of novel components such as geodesic masking and feature interaction via cross-attention in enhancing learning; and 3) the superior performance and efficiency of our proposed methods in challenging segmentation and classification tasks.
\end{abstract}

\section{Introduction}
\label{sec:intro}
In recent years, geometric deep learning has significantly advanced shape analysis and computer graphics applications, such as semantic segmentation, shape correspondence, and shape classification \cite{hanocka2019meshcnn, farazi2023anisotropic, sharp2022diffusionnet, hu2022subdivision, feng2019meshnet, fey2018splinecnn, bronstein2017geometric, monti2017geometric}. With the recent success of Vision Transformers (ViT) in image classification and segmentation \cite{dosovitskiy2020image,liu2021swin}, various methods like those in \cite{zhao2021point,liang2022meshmae, chen2022structure, yun2019graph} have adopted different transformers, including BERT-style, for unstructured data representation in graphs, 3D meshes, and point clouds, claiming to provide superior global and local feature learning through self-attention. Unlike vision transformers for images, unstructured 3D meshes and point clouds present challenges in tokenization, structural encoding, and generalizability. The primary obstacle is the absence of a canonical ordering and the inherent variability in size and geometry.

\subsection{Patch-based Transformer}Most geometric transformers utilize self-attention mechanisms that rely on node tokens \cite{dwivedi2020generalization,guo2021pct, yun2019graph, chen2022structure}. This is mostly because graph and point cloud datasets have a small number of nodes. For 3D meshes with a large number of nodes, the node and edge-based tokenization are computationally intractable as its computational cost based on global self-attention is $\mathcal{O}(n^2)$. We follow the sub-structure tokenization approach to solve this computational bottleneck in large 3D mesh polygons. Using patches over node tokens can alleviate the quadratic complexity of global all-to-all attention. Moreover, by combining learned node-level structural embeddings, the model ensures locality awareness while enabling global attention across all nodes in polygon meshes. In this study, we demonstrate that using a hybrid approach with spectral-preserving tokenization allows us to create a scalable geometric transformer framework without the need for sampling, remeshing, or size unification, making it adaptable to arbitrary mesh sizes.

\subsection{Tokenization} Generating patches for unstructured 3D meshes is more complex than for images with a uniform grid and size. 3D polygon meshes have varying numbers of nodes, and generating patches is less intuitive due to the lack of a consistent node ordering or structure. Additionally, partitioning the mesh into meaningful patches requires careful consideration of the geometric properties, making the process more challenging compared to the uniform grids used in images. This study proposes a spectral-preserving method inspired by the algebraic multigrid literature, known as \textbf{root-node selection} \cite{liu2019spectral}. We utilize this method to shape the partitions by employing a clustering algorithm that is based on an anisotropy-aware edge distance matrix calculated using Laplacian. 

Generally, a lack of geometrically meaningful structure encodings is present in similar studies on geometric transformers, specifically point clouds and 3D meshes. For the 3D mesh and point cloud, in \cite{liang2022meshmae, zhao2021point, wu2022point}, they employed the absolute positional encoding with 3D $xyz$ coordinates. However, as discussed in \cite{chu2021conditional}, a strong SE captures the relation and implicit distance among as many tokens as possible. Heat Kernel Signature (HKS) with a simple feed-forward network can possibly play as a candidate to encode relative positional information and serve as SE. 

\subsection{Dense Node-level Prediction}
A key challenge in using patch-based transformers for geometric data with varying input sizes arises in dense prediction tasks such as segmentation. Specifically, for semantic segmentation, node-level prediction is essential, yet translating features from patch-level to node-level is not a straightforward one-to-one process. To address this, we propose two approaches. First, we introduce a U-Net-type architecture that incorporates multiple patch-size embeddings, culminating in the summation of these embeddings to enhance node-level feature representation. Alternatively, we explore concatenating the learned features from the backbone with the final interpolated features. For this approach, we employ a cross-attention layer, which offers dual benefits. Firstly, it serves as a feature interaction module within the transformer layers during training. Secondly, it facilitates the aggregation of two inherently distinct feature representations. Overall, the transition from patch-level to node-level representation is crucial for node-level prediction tasks, and our strategies are designed to optimize this process effectively. 
 
Besides this model, we use a vanilla model with just point-wise MLPs without convolution and attention modules throughout all experiments. This vanilla model does not solely outperform other models but shows the significance of the SE and geodesic masking and proves the importance of GeoTransformer layers without solely being dependent on the backbone. This enlightens future studies on mesh and point cloud transformers. Overall, our goal is to develop various components of a patch-based mesh transformer specifically designed to handle inputs of arbitrary sizes. This model incorporates multiple geometrically-aware components optimized to enhance the processing of 3D meshes effectively. Also, we show how HKS serves as a strong explicit bias toward the isometry invariance of the model by encoding implicit structural and positional embedding.

\begin{figure*}[t]
 \centering
 \includegraphics[scale = 0.47,trim={0cm  0.3cm 0.0cm  0cm } ,clip]{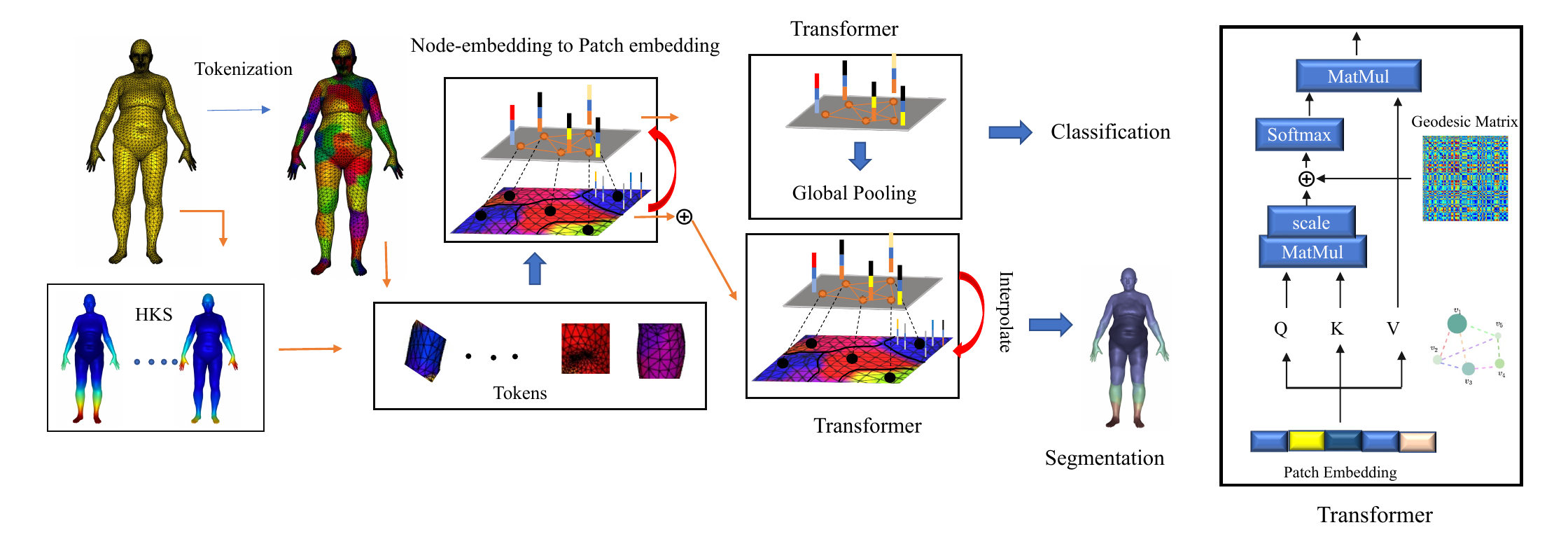}
 \caption{Our framework at a glance. The two-level approach can be extended to include multiple layers of varying resolutions, each with different patch sizes, which is a common practice for segmentation.}
 \label{fig:model}
 \vspace{-5mm}
\end{figure*}

\noindent\textbf{Contribution.}
In this paper, our contributions for the GeoTransformer are:
\begin{enumerate}
    \item We introduce a generalizable and scalable approach for 3D mesh transformers, leveraging geometric-aware components such as root-node aggregation tokenization and heat diffusion-based positional encoding. 
    \item By comparing a vanilla model with just point-wise MLPs, we show how a backbone specific to 3D meshes can further distill local spatial information for better performance. 
    \item We propose a geodesic attention masking for the transformer to attend to specific tokens. We will experimentally show if this approach helps the vanilla model or the DiffusionNet-backed framework. 
    \item We also provide an optional Multi-resolution approach for segmentation to mimic a U-NET-type architecture for dense segmentation. We further prove that using node-level feature learning would suffice to avoid multi-resolution transformers.
    \item We introduce a cross-attention module to serve as feature interaction with backbone and aggregation as the final step.
\end{enumerate}

\section{Related Works}
\subsection{Geometric Deep Learning on Mesh}

Geometric deep learning on meshes has emerged in different representation learning based on feature aggregation and learning paradigms using vertex, edge, or face features. Notable, frameworks like MeshCNN \cite{hanocka2019meshcnn} use edge-based networks equipped with downsampling and pooling for different shape analysis tasks. Frameworks based on feature learning on vertices \cite{vertex1,vertex2,vertex3,fey2018splinecnn}, use the neighborhood feature aggregations that usually are not robust to re-meshing and tend to overfit to mesh connectivity. Other vertex-based methods like \cite{sharp2022diffusionnet} apply diffusion kernels with varying size kernels to address the long-range dependency, making it a great global and local feature representation method. Face-based methods emphasize efficient and effective strategies for gathering information from adjacent faces. Study in \cite{hu2022subdivision} uses the fine-to-coarse hierarchical structure equipped with pooling and convolutions for different applications, including segmentation and shape correspondence. MeshNet \cite{feng2019meshnet} uses the adjacent face feature aggregation using two mesh convolution layers. In \cite{hertz2020deep} creates geometric textures, using $3$-face convolution with a subdivision-based upsampling strategy. Except for \cite{sharp2022diffusionnet}, these methods use the downsampling of meshes for pooling to capture high-range dependency. Furthermore, most methods are not robust to re-meshing and changes in mesh connectivity and require re-sampled mesh with lower resolution. 

\subsection{Geometric Transformers}
Since the original Transformer proposed \cite{vaswani2017attention}, researchers have adopted transformers for vision tasks, from classification to object detection. ViT and its variants \cite{dosovitskiy2020image,liu2021swin} used image patches with positional encoding to them as a series of words. To transfer such an approach for 3D mesh and point cloud structure, in \cite{liang2022meshmae}, they used the first patch-based approach with re-meshing the original mesh to fix the dimension of mesh and patches. They successfully adopted a self-supervised masked auto-encoder for pre-training on larger mesh datasets for segmentation and classification downstream tasks. In \cite{li2022meshformer}, they used Ricci-Flow-based clustering as a pre-segmentation step to use large patches similar to the ground truth equipped with a graph transformer. They tested their model on teeth segmentation and part segmentation tasks on general shapes. At the same time, several point-cloud-based approaches were proposed, like  \cite{zhao2021point,wu2022point}, which use sampled points from a mesh with a high computational cost for their global attention computation. Recently, \cite{wong2023heat} used a sampled meshed input and introduced fixed-size heat kernels to alleviate the global attention cost by restricting several computations within all pairs. In \cite{li2022laplacian}, they used a hybrid point cloud and mesh approach to induce geometry and topology-aware representation learning for various tasks, including part segmentation. The closest to our approach in terms of patch-based embedding is \cite{he2023generalization}, which is based on small graphs and different topologies. 

\section{Background and Overview}

This section elaborates on our proposed mesh transformer model for polygon meshes. Our model can be used for any type of mesh, including 3D triangle mesh, tetrahedral mesh, or any simplicial complex that can be represented with a geometric graph structure. We will delve into the details of each component and present a corresponding ablation study for each to show the merits of the proposed methodology. Specifically, we refer to the vanilla model without any mesh-based neural network backbone for node-level feature learning by simply using MLPs. The overall architecture is illustrated in ~Fig.\ref{fig:model}. We start with a preliminary overview to introduce the notation and geometric definitions.   
\subsection{Notation}
In this paper, we represent the 3D mesh as $M = (\mathit{V},\mathit{E})$, in which $V$ and $E$ represent vertices and edge-set of the mesh equipped with its corresponding Laplace Beltrami Operator (LBO) $L$. As we use the LBO in backbones and computation of the HKS features for structural encoding, we use the tuple $(\mathit{\Phi},\mathit{\Lambda})$ for the eigenfunctions and eigenvalues matrices of LBO. The mesh consists of $N = |V|$ nodes and $E = |E|$ edges and $|P|$ distinct patches after partitioning the mesh. Using a geodesic distance matrix for patch-level positional masking, we represent it with $G \in \mathbb{R}^{P \times P}$. Unless stated otherwise, node-level attributes are denoted by their associated patch numbers as subscripts. Layer-level attributes are indicated with superscripts $()^{layer}_{node, patch}$, like $x^l_{i,p}$ is the node feature of node $i$ in patch $p$ at layer $l$. Mesh partitions and their attributes at transformer layers are represented with a single subscript, like $M_i = (\mathit{V_i},\mathit{E_i})$. 

\textbf{definition}
\label{def:hks}
Given a point \(x\) on the mesh \(M\), we define the Heat Kernel Signature (HKS) as a function in a temporal domain: \(HKS(x): \mathbb{R}^+ \rightarrow \mathbb{R}\), \(HKS(x, t) = k_t(x, x)\) in which 
$k_t(x, x) = \sum_{i=0}^{\infty} e^{-\lambda_i t} \phi_i(x)^2 $
Here $t$ represents different time scales, and $\phi$ is the eigenfunction of LBO. 

\begin{figure}
 \centering
 \includegraphics[scale = 0.4,trim={0.0cm  -0.0cm 0.0cm  -0.2cm } ,clip]{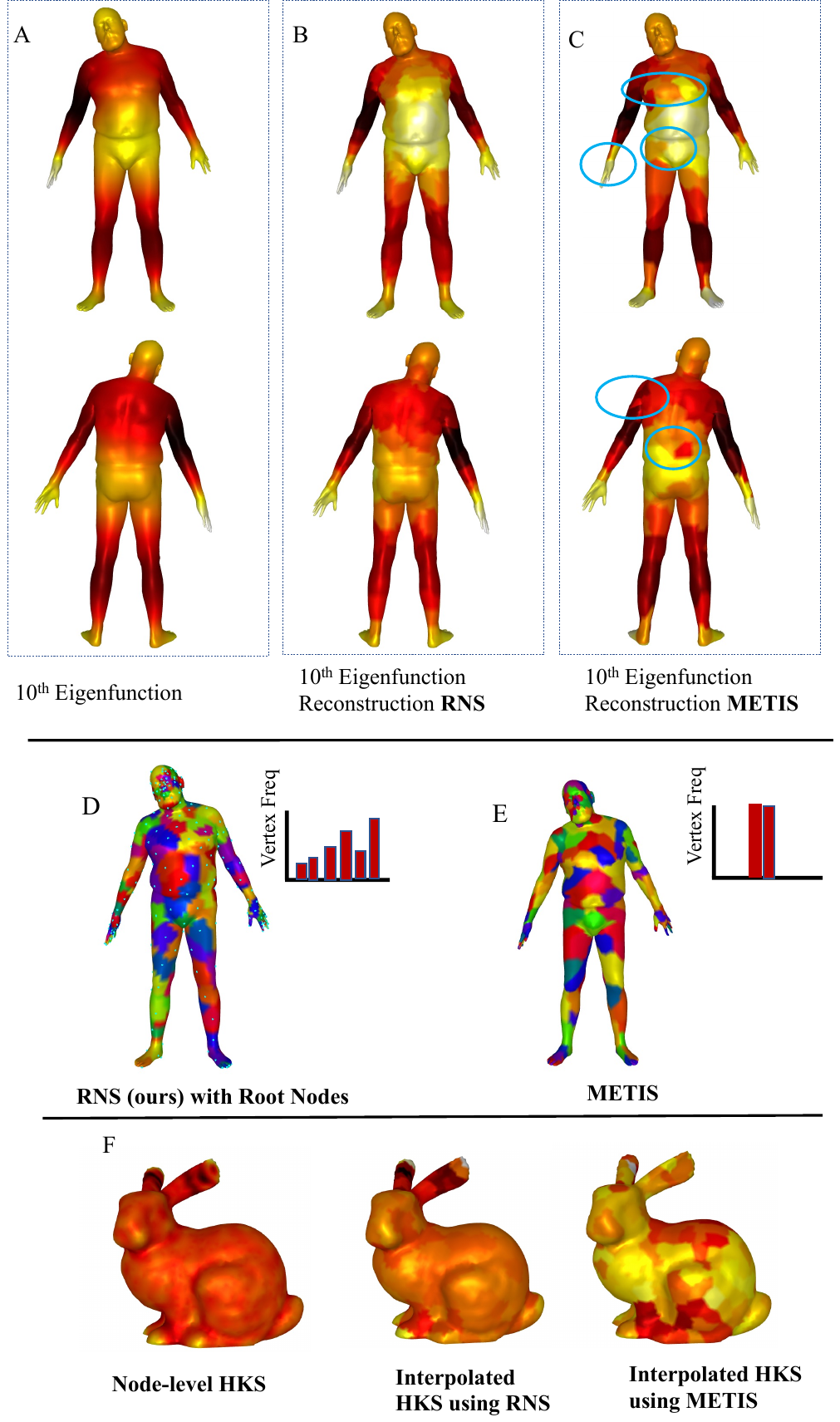}
 \caption{Comparison between two tokenization methods, Root-Node Selection (RNS) and METIS, reveals significant differences. On the left (A-C), the interpolated approximation of LBO eigenfunctions using RNS demonstrates superior performance compared to METIS. Additionally, on the bottom right of the bunny, the Heat Kernel Signature (HKS) used for patches outperforms the METIS partitioning method. To visually demonstrate the supernodes and partitions based on RNS, Figure $D$ showcases the results on a sample object alongside its counterpart method, METIS ($E$).}
 \label{fig:patch_comp}
 \vskip -0.15in
\end{figure}

\subsection{Mesh Paritioning}
\subsubsection{Why Root-Node Selection}
Patch-based tokenization techniques have been seldom employed in 3D meshes and graphs. This is partly because tokenizing at the node level typically does not introduce additional computational overhead. This advantage is particularly relevant for graphs and point clouds, which either naturally contain fewer nodes or are commonly downsampled to lower resolutions, as is often the case with point clouds and 3D mesh transformers. In 3D meshes, models like in \cite{liang2022meshmae} require fixed input size tokens, and therefore, re-meshing and unifying the input size is mandatory. We introduce the root-node selection and partitioning approach that enjoys spectral preserving and anisotropy-aware properties \cite{liu2019spectral}. Also, given both root nodes (super-nodes) and clustering assignments based on the closest point strategy, we can introduce a set of desired strategies for a general framework. One is to define a geodesic distance matrix based on just super-nodes optionally.
Furthermore, we will use this for the geodesic attention mask. Given the cluster assignment matrix, one can approximate various geometric operators like Laplacian as introduced in \cite{liu2019spectral}. This is more generalizable with datasets with significantly varying size inputs. Another merit of this technique is that patches have different sizes based on the underlying geometry. A good illustration is depicted in Fig~\ref{fig:patch_comp}. 

As shown in Fig~\ref{fig:patch_comp}, we also compare the performance based on METIS \cite{karypis1998metis}, a well-known graph partitioning method that is very efficient in creating balanced size patches on a graph and a 3D mesh. The difference is that METIS is not designed specifically for 3D meshes. Consequently, it lacks sensitivity to the underlying geometry. As demonstrated, it fails to preserve the spectrum when approximating the Heat Kernel Signature (HKS) on partitions and Laplacian-Beltrami Operator (LBO) eigenfunctions.

\subsubsection{Notation} Using any partitioning algorithm, we denote the partitioned sub-meshes of $M$ as  $\{ m_1, \ldots, m_P \}$ such that  $M = \{m_1 \cup \ldots \cup m_P\} \text{ and } m_i \cap m_j = \emptyset, \, \forall i \neq j$. 
\subsubsection{Root-Node Selection}
Given the Laplacian \( L \in \mathbb{R}^{n \times n} \) and the mass-matrix \( W \in \mathbb{R}^{n \times n} \), the goal is to find the root-nodes and its associating clusters by using closest point strategy. Given a graph with \( n \) nodes represented by the nonzeros of the matrix \( L \) interpreted as undirected edges, the objective is to perform node clustering and root-node selection while considering the diffusion of information and the associated node weights according to \( L \) and \( W \), respectively \cite{liu2019spectral}. The edge distance for a 3D triangle mesh can be defined with

\begin{equation}
    D_{ij} = \max \left(\frac{{(W_{ii} + W_{jj)}}^{0.5}}{-L_{ij}} , 0\right)
\end{equation}
Using a symmetric matrix of edge distances, we perform \textbf{k-medoid clustering} of the graph nodes based on shortest path distances \cite{struyf1997}. This is accomplished efficiently using a modified Bellman-Ford method and Lloyd aggregation method proposed by Bell \cite{bell2008}. 

\subsection{Node and Patch-Level Embeddings}
Unlike previous ViT-style transformers that work with fixed-size patches, i.e., the number of nodes being fixed per patch, we propose using patch embeddings without using techniques like padding as inputs can drastically vary in terms of the number of nodes. We eliminate the cumbersome need to re-mesh and re-sample all meshes to achieve fixed-size inputs for operation. Therefore, we need a feature representation and aggregation scheme within each patch to circumvent the common concatenation of node features in fixed-size geometry. We consider two node-level feature learning strategies. 

\textbf{GeoTransformer with MLP (vanilla)} The first option is to use point-wise MLPs, which do not include convolution and local awareness and are very efficient in computation. Such a simple approach, equipped with structural encoding and geodesic masking, has two advantages. First, it shows how such a naive model without any backbones, like in \cite{sharp2022diffusionnet, qi2017pointnet++, hanocka2019meshcnn}, can perform well compared to some models with significantly higher computational cost. Second, it can perfectly serve as a baseline for the model with a DiffusionNet backbone and all the ablation studies.

\textbf{GeoTransformer with a Backbone} Second is the Geotransformer with a backbone, i.e. equipped with a spatial prior module, that can satisfy a few criteria: $\textbf{1})$ learning spatial and local features for patch embedding, $\textbf{2})$ being robust to re-meshing $\textbf{3})$ being scalable to large meshes with linear computation cost. Given these criteria, DiffusionNet \cite{sharp2022diffusionnet} is the most suitable candidate that learns features based on varying kernel support. 

\subsubsection{Backbone Architecture}
We will begin by explaining the chosen backbone, DiffusionNet layers. With input node features that are linearly projected to higher dimension $x_i \in \mathbb{R}^{1\times D}$ \cite{sharp2022diffusionnet}:

\begin{equation}
    h_t(x) := \Phi \begin{pmatrix} e^{-\lambda_0 t} \\ e^{-\lambda_1 t} \\ \vdots \end{pmatrix} \odot \left( \Phi^T \bar{W} x \right)
\end{equation}
Here $\bar{W}$ is the mass matrix, $\odot$ is the Hadamard product, and the eigenvectors $\boldsymbol{\phi}_i \in \mathbb{R}^V$ are solutions to:
\[ \mathbf{L}\boldsymbol{\phi}_i = \lambda_i \mathbf{\bar{W}}\boldsymbol{\phi}_i, \]
This simple diffusion layer learns time $t$ per channel. With a sufficient number of consequent layers, the network can learn varying-size kernels for better information propagation. Equipped with MLP and optional spatial gradient features, it can create anisotropic filters to learn robust features better. If we apply a series of $L$ diffusion layers equipped with optionally HKS input node features, we can express the updated node features within each patch as:

\begin{equation}
\label{eq2}
     x_{i,\ell+1}^p = f_{\theta}([h_t(x_{i,1}^p)_{i,1}^p, x_{i,\ell}^p, g_{i}^p]) + x_{i,\ell}^p
\end{equation}
For our vanilla model with just point-wise MLP without using any GNN or DiffusionNet node-wise learning we have: 
\begin{equation}
\label{mlp_eq}
     x_{i,\ell+1}^p = f_{\theta}([ x_{i,\ell}^p]) + x_{i,\ell}^p
\end{equation}
Here $f_{\theta}$ is MLP layers with proper normalization and activation functions, and $g$ serves as optional spatial gradient features. Next, we have to learn an embedding based on the node-level features to form a $x_p \in \mathbb{R}^{1\times D}$ patch embedding of the mesh.  

\subsubsection{Aggregation of Features}\noindent The final step is to aggregate the node features in each $m_p$ such that:
\begin{equation}
    x^p = \sum_{i \in V_p} {Softmax(x_{i,L}^p). h_{\theta}(x_{i,L}^p)}
\end{equation}
Here $x_p$ is the fixed-size embedding for patch $p$. Alternatively, a simple average pooling is viable. However, it is not sensitive to the importance of nodes in each patch. The proposed aggregation allows the model to learn a weighted average to embed the patch-level features better.

\subsection{Structural Embedding}
In our network, we employ the Heat Kernel Signature (HKS) as a relative positional embedding. The HKS, integrated with a feed-forward network, brings rich geometric properties to our GeoTransformer. This method, characterized as a weighted sum of the squares of eigenfunctions, is not affected by the order or signs of these eigenfunctions, enhancing its robustness to perturbations and ensuring isometry invariance. Furthermore, in contrast to the vanilla model that uses Multi-Layer Perceptrons (MLPs) and lacks sensitivity to local geometry, including HKS and self-attention, it is crucial for effectively capturing global geometric features. It is noteworthy to mention that to get patch-level HKS, we simply average the values of each node per patch to get patch-level SE.






\begin{figure}[htbp]
  \centering
  \begin{minipage}[b]{0.9\linewidth}
    \centering
    \includegraphics[scale=0.49,trim={0.5cm 0cm 0.0cm 0.3cm},clip]{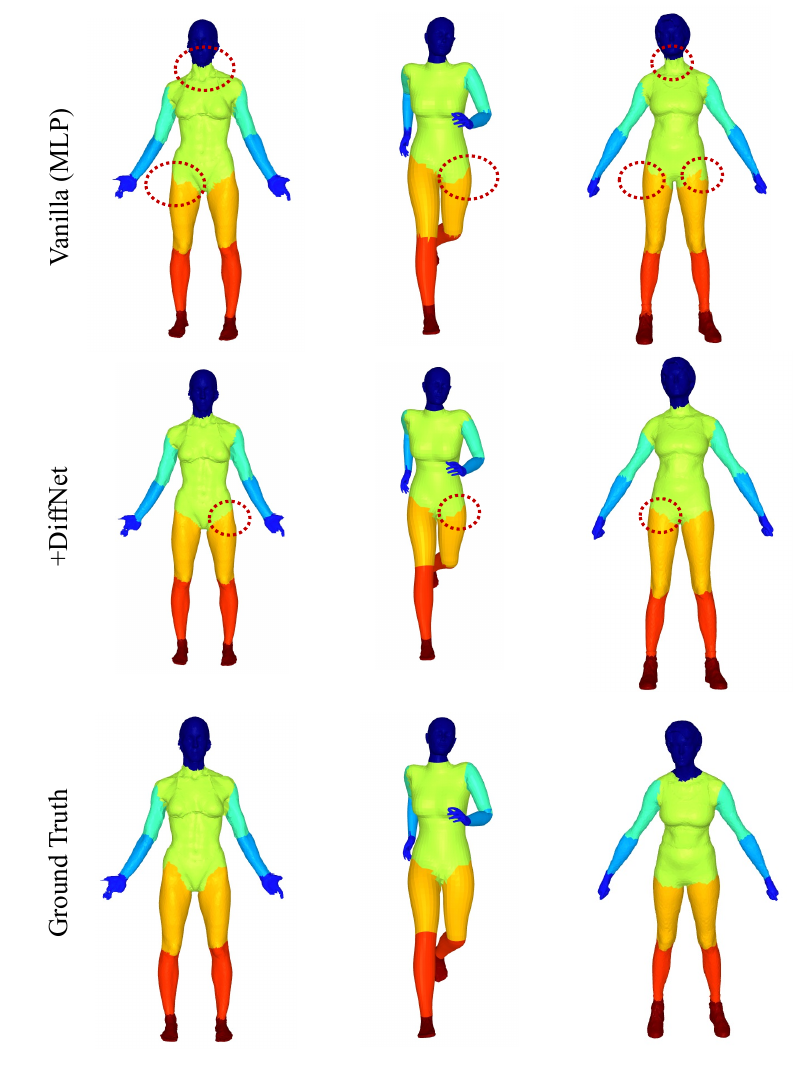}
    \caption{Segmentation results on human body segmentation.}
    \label{fig:human}
  \end{minipage}
  
  \vfill 

  \begin{minipage}[b]{0.9\linewidth}
    \centering
    \includegraphics[scale=0.4,trim={0.0cm 0.0cm 0.0cm -0.3cm},clip]{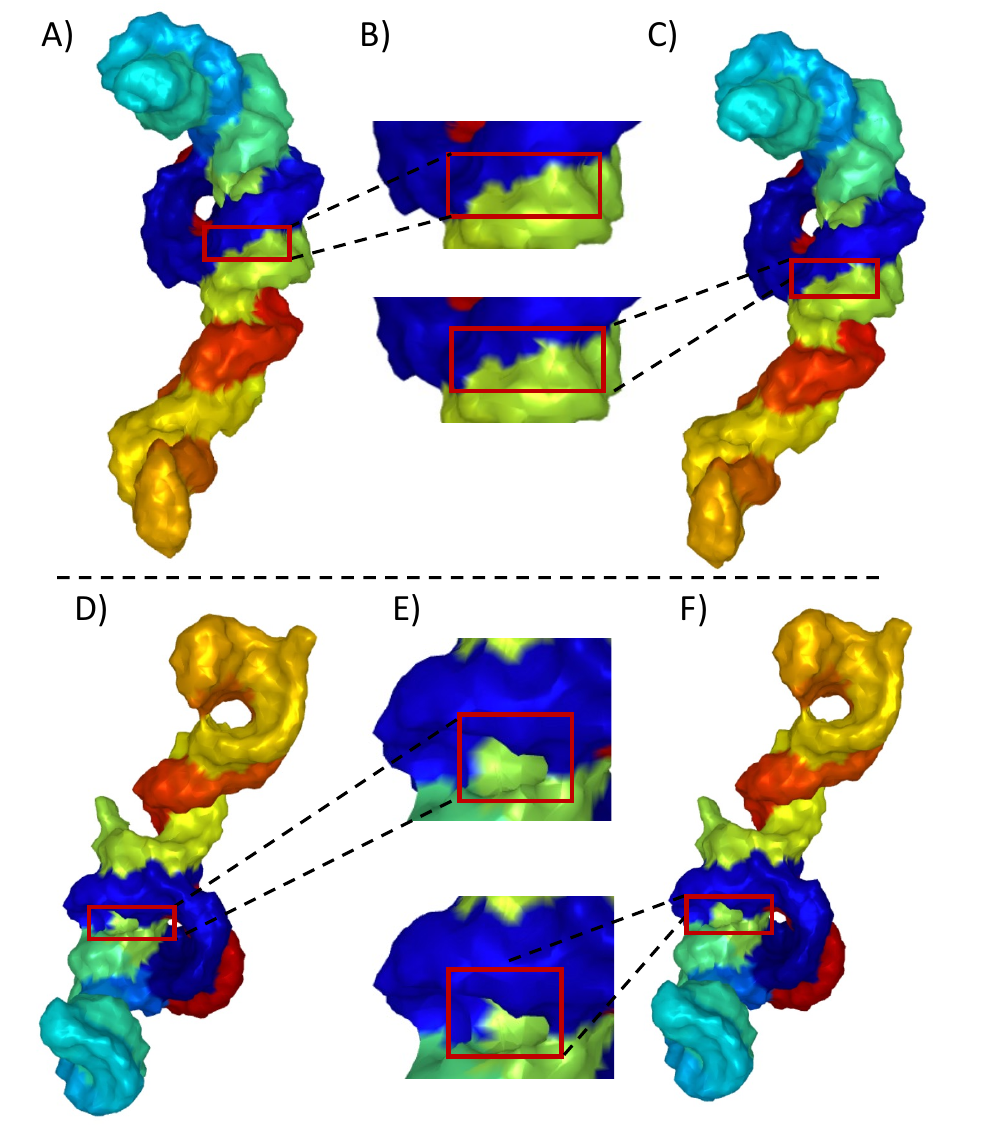}
    \caption{Segmentation results for RNA mesh segmentation. $A$ and $D$ are the ground truth. $C$ and $F$ are the results.}
    \label{fig:rna_segs}
  \end{minipage}
\end{figure}



\subsection{Transformer Layer}

We use the original transformer model \cite{vaswani2017attention} with optional masked structural attention using a geodesic distance matrix. We use the average of HKS per node to get patch-level SE. Each transformer layer has two segments: a Multi-Head Self-attention (MHA) module and a feed-forward network (FFN). Let the learned patch embeddings be $X_p = x_{p_1}, \ldots, x_{p_n} \in \mathbb{R}^{P \times D}$ denote the input of the MHA module, where $D$ is the dimension of the hidden layer, and $P$ is the number of patches. Then $X_p$ is projected by $W_Q \in \mathbb{R}^{D \times d_K}$, $W_K \in \mathbb{R}^{D \times d_K}$, and $W_V \in \mathbb{R}^{D \times d_V}$ to $Q, K, V$. The self-attention is then calculated as \cite{ying2021transformers,vaswani2017attention}:
\begin{equation}
    Q = X_p W_Q, \quad K = X_pW_K, \quad V = X_pW_V
\end{equation}
\begin{equation}
     \text{Attention}(X_p) = \text{softmax}\left(A_{ij} + \frac{QK^T}{\sqrt{d_k}}\right)V 
\end{equation}
Here, $A_{ij} \in \mathbb{R}^{D \times D} $ is an attention mask based on the optional geodesic distance matrix thresholded to attend to specific nodes. This term can be task-specific with a hyper-parameter to threshold the geodesic distance matrix. We will see in our ablation studies that HKS suffices, but a structural mask like a geodesic distance matrix, which is novel to this study, can improve the performance. Basically, if we choose not to attend to patches of a specific range, we assign $-\infty$ and $1$ otherwise. Since we already have the super-nodes, we calculate the geodesic distance among them to build the attention mask matrix. This is much cheaper than computing all pairs of geodesic distance computation. In Fig ~\ref{fig:model}, you see the geodesic distance matrix added before Softmax on the right side of the transformer architecture. Using this attention masking strategy and structural encoding in patch-level embedding is more crucial than using the vanilla model with simple MLPs, as they lack spatial and local awareness. Moreover, such masking can reduce the computational cost for large-scale meshes as well.



\textbf{Implementation of Transformer}
For implementation, we follow the \cite{ying2021transformers} but do not include any further 
structural encoding. We formally represent the transformer layer as below:
\begin{equation}
X^{(l)} = \text{MHA}\left(\text{LN}(X^{(l-1)})\right) + X^{(l-1)} \in \mathbb{R}^{P \times D}  \tag{8}
\end{equation}
\vspace{-2ex}
\begin{equation}
X^{(l)} = f_{\theta}\left(\text{LN}(X^{(l)})\right) + X^{(l)}  \in \mathbb{R}^{P \times D} \tag{9}
\end{equation}

$f_{\theta}$ is a simple MLP layer, and LN is Layer Normalization. Given the specific task, classification, or segmentation, we use global average pooling or upsampling to dense node prediction. 

\subsection{Feature Interaction and Aggregation Module}
Initially, we employ patch-level and node-level feature representation learning, and then the GeoTransformer layers operate on patch-level embeddings. As a result, the interaction of the spatial learning module, either the vanilla model or the DiffusionNet backbone, becomes restricted. Thus, we use a cross-attention layer for both better interaction and also the aggregation of features from dense node-level and patch-level, especially for segmentation tasks. In this cross-attention, queries come from the GeoTransformer's output, and key/values come directly from the node-level feature learning layers. 
\begin{equation}
F^{\text{out}} = F + \lambda \cdot \text{CrossAtt}\left(\text{norm}(x), \text{norm}(X)\right),
\end{equation}
\vspace{-2ex} 
\begin{equation}
F^{\text{out}} = \text{FFN}(\text{norm}(F^{\text{out}}))
\end{equation}

where \(F\) represents the last GeoTransformer layer, \(\lambda\) represents the leaning parameter to control the importance of cross-attention output, $FFN$ is a feed-forward network, and \(norm\) is layer normalization function. 

\section{Experiments and Analysis}

 \begin{table}
\caption{Results on Human Part Segmentation. $\ast$ shows the results without using majority voting for a fair comparison with ours and other methods that have not used this. $\ast\ast$ shows the results based on the same configuration we had in our experiment for DiffusionNet.}
\label{human_t}

\begin{center}
\renewcommand{\arraystretch}{1.0}
\begin{small}
\begin{tabular}{lcr}

\toprule
Method & Input & Accuracy  \\
\midrule

PointNet++ \cite{qi2017pointnet++} & Point & 90.8\% \\
DGCNN \cite{wang2019dynamic} & point & 89.7\% \\
ACNN \cite{boscaini2016learning} & Mesh & 83.7\% \\
HSN \cite{wiersma2020cnns}  & Mesh & 91.1\% \\
MeshWalker \cite{lahav2020meshwalker} & Mesh & \textbf{92.6\%} \\
CGConv \cite{yang2021continuous} & Mesh & 89.9\% \\

Laplacian2Mesh \cite{dong2023laplacian2mesh} & Mesh & 88.6\% \\

Mesh-MLP \cite{dong2023mesh}  & Mesh & 90.6\% \\
MeshMAE  \cite{liang2022meshmae}  & Mesh & 90.2\% \\
SubDivNet$^{\ast}$ \cite{hu2022subdivision} & Mesh & 91.1 \\
DiffusionNet-hks$^{\ast\ast}$  \cite{sharp2022diffusionnet}  & Mesh & 91.5\% \\
\midrule
Ours (MLP)  & Mesh & 90.3\% \\
Ours (MLP MultiRes)  & Mesh & 90.7\% \\
Ours (DiffNet)  & Mesh & \textbf{92.6\%} \\
\bottomrule
\end{tabular}
\end{small}
\end{center}
\end{table}

\subsection{Setup}
We use the same setting in DiffusionNet to compare our architecture and training procedure for all tasks if using it as the backbone. We use the hidden layer feature dimension of $128$ for all tasks except for RNA segmentation for the vanilla model with the dimension of $256$. All inputs are centered and scaled to a unit sphere. HKS are sampled at $16$ values of \(t\) logarithmically spaced on \([0.01, 1]\). We do not use any data augmentation as our network is invariant to isometry. Since we used the same settings in DiffusionNet, we re-ran the experiments and reported the number for DiffusionNet. ADAM optimizer is used for all the tasks with a learning rate of $0.001$ and weight decay of $0.5$ after each $50$ epoch. Experiments with DiffusionNet all included $4$ layers, and the number of transformer layers is $2$. For the vanilla model, we use $4$ MLP layers following DropOut with the probability of $0.5$ and LayerNorm. The following individual task setup will provide more specific details for each task.

\begin{table}[htbp]
\caption{Ablation results for segmentation tasks are presented in this table. To understand how to interpret the table: an entry is marked with a check if HKS was used as the structural embedding (SE), or if masking was applied in the experiment. 'RNS' stands for root-node-selection, and 'MET' refers to METIS partitioning. 'HB' and 'RNA' indicate human body segmentation and RNA segmentation tasks, respectively.}
\label{ablation_t}
\centering
\renewcommand{\arraystretch}{1.1} 
\begin{scriptsize} 
\begin{tabular}{lcccccr}
\toprule
Method & Mask. & RNS/MET & Part. & HKS & Task & Acc. \\
\midrule
Diff & \checkmark & RNS          & 256  & \checkmark & HB & 92.6\% \\
Vanilla & \checkmark & RNS          & 256 & \checkmark & HB & 90.7\% \\
Vanilla & \checkmark & RNS          & 256    & \checkmark & RNA & 85.5\% \\
Diff &            & RNS         & 256  & \checkmark & HB & 92.6\% \\
Diff &            & RNS          & 256    & \checkmark & RNA & 91.1\% \\

\midrule
Vanilla &            & RNS          & 256  & \checkmark & HB & 82.0\% \\
Vanilla &            & RNS          & 256  & \checkmark & HB & 89.4\% \\
\midrule
Diff &            & RNS          & 128-512 & \checkmark & RNA & 91.4\% \\
Vanilla &            & RNS          & 256    & \checkmark & RNA & 85.0\% \\
\midrule
Vanilla &            & MET          & 256    & \checkmark & RNA & 85.1\% \\
Vanilla &            & MET          & 256  & \checkmark & HB & 90.5\% \\
\midrule
Vanilla &            & RNS          & 256    &            & RNA & 84.5\% \\
Vanilla &            & RNS          & 256  &            & HB & 87.5\% \\

\bottomrule
\end{tabular}
\end{scriptsize}
\end{table}

\subsection{Implementation Details}

By default, we used $256$ partitions for each mesh using root-node selection, hence patches. Also, for the vanilla model, we also included METIS for the ablation study. For the segmentation task, we used two levels with just $256$ patches; however, for more sophisticated networks, we optionally used UNET-type with multi-resolution for the MLP-only model. For the RNA segmentation task, we used two settings for partitioning, $128-512$ (multi-res) and $256$, as the complexity of shape and the number of labels for segmentation are potential contributing factors. This is shown as an ablation study, and it is reported in the results section. Precomputation of LBO eigendecomposition and gradient features were all done using a $16$ core CPU computer with one RTX3090 GPU for training the models. We used Pytorch\cite{paszke2019pytorch}, and  PyG \cite{fey2019fast} for our training in Python. 



\subsection{Segmentation Experiments }
\textit{Molecular Segmentation}
In our first experiment for segmentation, we used a challenging dataset with many segmentation parts for a complex mesh structure. We use the RNA molecules dataset introduced by  ~\cite{poulenard2019} for segmenting them to $259$ functional components. This dataset comprises $640$ RNA mesh objects with around $15k$ vertices, pulled from the Protein Data Bank~\cite{berman2000protein}. Labels are per-vertex according to different atomic categories with a random $80/20$ train-test split. We use a $4$ DiffusionNet layer for the main task, and for the vanilla model, we use $4$ MLP layers, including normalization and dropout layers. We re-ran the code for DiffusionNet for fair comparison as the exact hyper-parameters and settings are not defined. For transformer layers, we use $2$ layers with no dropout. We also compared our results to the baseline in \cite{atzmon2018pointconvolutional,fey2018splinecnn,kostrikov2018surface, poulenard2019} in Table ~\ref{rna_t}. The illustration of results is also shown in Fig ~\ref{fig:rna_segs}. 

\begin{table}
\caption{Results on RNA Segmentation. $\ast$ denotes results after running with the same configuration as in our experiment and the one reported in \cite{sharp2022diffusionnet}.}
\label{rna_t}
\begin{center}
\renewcommand{\arraystretch}{1.0}
\begin{small}
\begin{tabular}{lcr}

\toprule
Method & Input & Accuracy  \\
\midrule

PCNN \cite{atzmon2018pointconvolutional} & Point & 78.0\% \\
SPHNet \cite{poulenard2019} & Point & 80.1\% \\  
SplineCNN \cite{fey2018splinecnn} & Mesh & 53.6\% \\
SurfNet \cite{kostrikov2018surface} & Mesh & 88.5\% \\
DiffusionNet$\ast$ \cite{sharp2022diffusionnet} & Mesh & 90.1\% \\
\midrule
Ours (MLP)  & Mesh & 85.5\%\\
Ours (DiffNet)  & Mesh & \textbf{91.4}\%\\
\bottomrule
\end{tabular}
\end{small}
\end{center}

\end{table}

\subsection{Human Body Segmentation}
The human body dataset initially was labeled by \cite{Maronetal2017}, including $381$ training mesh objects from SCAPE \cite{Anguelovetal2005}, FAUST \cite{Bogoetal2014}, MIT \cite{Vlasicetal2008}, Adobe Fuse \cite{AdobeMixamo2021}, and $18$ test objects from SHREC07 \cite{Giorgi2007SHREC} dataset. The task is to segment human bodies into $8$ different semantic parts. Here, we do not use the simplified or re-meshed shapes like in \cite{hu2022subdivision, hanocka2019meshcnn, liang2022meshmae,lahav2020meshwalker}. As indicated in Table ~\ref{rna_t}, the vanilla model reaches $90.7\%$, and with DiffusionNet backbone, it reaches $92.6 \%$. In this experiment, we used $256$ partitions for both methods. We used the Geodesic masking for the MLP model for this experiment, which we further used for the ablation study. Our vanilla model surprisingly performed well and was superior to the closest to our work \cite{liang2022meshmae}. This further reinforces the value of structural encoding. For the visualization results, please see Fig ~\ref{fig:human}.

\subsubsection{Classification Experiments }
Manifold40 \cite{hu2022subdivision,Wu2015}, containing 12,311 shapes in 40 categories, is a widely used benchmark for 3D shape classification. We follow \cite{hu2022subdivision} for using the meshed version without non-watertight mesh. The results are in Table ~\ref{Manifold40}. The MLP model, which is the most efficient among all these models, performs relatively well; however, the overall accuracy of DiffusionNet and our model is slightly worse than that of the state-of-the-art model. One possible reason behind this is that DiffusionNet does not support multi-batching, making the training on many data more challenging and unstable.  


\section{Ablation Studies}

\begin{table}
\captionsetup{skip=-4pt} 
\caption{Results on Manifold40 classification task.}
\label{Manifold40}

\begin{center}
\renewcommand{\arraystretch}{1.0}
\begin{small}
\begin{tabular}{lcccr}

\toprule
Method & Input & Accuracy  \\
\midrule

MeshNet\cite{feng2019meshnet}  & Mesh & 88.4\% \\

SubdivNet\cite{hu2022subdivision}  & Mesh & \textbf{91.2}\% \\
MeshMAE\cite{liang2022meshmae}  & Mesh & 91.1\% \\
Laplacian2Mesh\cite{dong2023laplacian2mesh} & Mesh & 90.9\% \\
DiffusionNet & Mesh & 88.3\% \\
\midrule
Ours (MLP) & Mesh & 89.2\%\\
Ours (DiffNet) & Mesh & 90.5\%\\
\bottomrule
\end{tabular}
\end{small}
\end{center}
\end{table}

\begin{table}
\captionsetup{skip=-4pt} 
\caption{Ablation study on the effectiveness of GeoTransformer in different scenarios. DiffNet: DiffusionNet layers, Diff: using DiffusionNet as the backbone. GeoTrans: GeoTransformer}
\label{ab2}

\begin{center}
\renewcommand{\arraystretch}{1.0}
\begin{small}
\begin{tabular}{lcccr}

\toprule
Method & DiffNet Layers & Accuracy  \\
\midrule

DiffusionNet  & 1 & 89.3\% \\
DiffusionNet  & 4 & 91.5\% \\
GeoTrans + Diff  & 1 & 91.2\% \\
GeoTrans + Diff  & 4 & 92.6\% \\
GeoTrans + MLP  & 0 & 90.7\% \\

\bottomrule
\end{tabular}
\end{small}
\end{center}
\end{table}

This section analyzes the importance of each component that has been proposed so far, particularly versus a vanilla model. We also show the results for partitions of different sizes in the RNA segmentation task. To better understand the Table ~\ref{ablation_t}. Some settings are fixed between each section of the table, like geodesic masking or using METIS for partitioning. We can infer that, for instance, geodesic masking specifically works best for the vanilla model as it needs more local feature learning focus. Using a threshold to limit the attention to neighboring patches enhances the performance accordingly. However, using the backbone does not change the performance noticeably. 

From the ablation results in Table ~\ref{ab2}, we can infer that the learning capacity is not solely based on the backbone itself. For instance, with just one layer of DiffusionNet, the performance is at $89.3\%$, whereas, when used as the backbone for GeoTransformer, the gap is at its maximum at $2\%$. Furthermore, when just using MLP and no spatial prior module, i.e., a backbone like DiffusionNet, the performance already passes $90\%$, which is very significant as this is computationally considerably cheaper than any other models. Another important insight is the role of geodesic masking when using the vanilla model.   
\section{Conclusion}
In this study, we proposed a geometry-aware 3D mesh transformer and addressed tokenization based on root-node selections. We further investigated the role of patch embedding and how heat diffusion-based learning modules and heat kernel signatures can serve as components of a mesh transformer. Last but not least, by comparing the proposed framework with a vanilla model with just point-wise MLP, we could scrutinize the role of different components, including the proposed geodesic attention mask, to control the locality and computation efficiency of the transformer block.


\end{document}